# Predição de Incidência de Lesão por Pressão em Pacientes de UTI usando Aprendizado de Máquina

Henrique P. Silva[1,2], Arthur D. Reys[2], Daniel S. Severo[2], Dominique H. Ruther[2], Flávio A. O. B. Silva[3], Maria C. S. S. Guimarães[3], Roberto Z. A. Pinto[3], Saulo D. S. Pedro[2], Túlio P. Navarro[4], Danilo Silva[1]

[1]*Universidade Federal de Santa Catarina, Florianópolis, SC*

[2]*Grupo 3778, Belo Horizonte, MG*

[3]*Rede Mater Dei de Saúde, Belo Horizonte, MG*

[4]*Universidade Federal de Minas Gerais, Belo Horizonte, MG*

{henrique.silva, arthur.reys, severo, dominique.ruther, saulo.pedro}@3778.care, flavio.bitar@materdei.com, {clarasalomaoguimaraes, zambelliortop, tulio.navarro}@gmail.com, danilo.silva@ufsc.br

***Abstract.*** *Pressure ulcers have high prevalence in ICU patients but are preventable if identified in initial stages. In practice, the Braden scale is used to classify high-risk patients. This paper investigates the use of machine learning in electronic health records data for this task, by using data available in MIMIC-III v1.4. Two main contributions are made: a new approach for evaluating models that considers all predictions made during a stay, and a new training method for the machine learning models. The results show a superior performance in comparison to the state of the art; moreover, all models surpass the Braden scale in every operating point in the precision-recall curve.*

***Resumo.*** *Lesões por pressão possuem alta prevalência em pacientes de UTI e são preveníveis ao serem identificadas em estágios iniciais. Na prática utiliza-se a escala de Braden para classificação de pacientes em risco. Este artigo investiga o uso de aprendizado de máquina em dados de registros eletrônicos para este fim, a partir da base de dados MIMIC-III v1.4. São feitas duas contribuições principais: uma nova abordagem para a avaliação dos modelos e da escala de Braden levando em conta todas as predições feitas ao longo das internações, e um novo método de treinamento para os modelos de aprendizado de máquina. Os resultados obtidos superam o estado da arte e verifica-se que os modelos superam significativamente a escala de Braden em todos os pontos de operação da curva de precisão por sensibilidade.*

**Palavras-chave:** Lesão por Pressão; Aprendizado de Máquina; Registros Eletrônicos.

## INTRODUÇÃO

Lesões por pressão (LP) são ferimentos na pele que podem se estender até tecidos subcutâneos sendo geralmente causadas por pressão prolongada em locais de protuberância óssea. Pacientes de UTI possuem risco elevado de desenvolver LP, com incidências reportadas de 3% a 33%, devido à relação de LP com fatores comuns a pacientes de UTI como baixa mobilidade, comorbidades, baixa perfusão, uso de vasopressores, entre outros fatores [1].

Segundo [2], LP são classificadas em 8 categorias, dentre as quais 4 estágios de progressão. Em geral, o tempo de recuperação, assim como a gravidade de danos físicos e psicológicos, aumentam com o estágio da LP. Além disso, tratar uma LP em estágio avançado tem custo muito superior que a prevenção destas lesões. Portanto, evitar a progressão dessas lesões é de grande importância. A prevenção é feita a partir de medidas que incluem frequente mudança de posição do paciente, uso de colchões especiais, uso de espumas multi-camadas, entre outras. No entanto, essas medidas exigem esforço elevado da equipe de enfermagem, bem como um custo elevado de materiais e processos. Nesse sentido, é inviável aplicar essas intervenções em todos os pacientes internados, sendo geralmente aplicadas em pacientes classificados como em risco a partir da escala de Braden, proposta em 1988 [3].

A escala de Braden consiste em avaliar seis fatores de risco sendo eles percepção sensorial, mobilidade, nutrição, fricção, umidade e atividade física. Existem muitos trabalhos que avaliam o poder preditivo da escala de Braden [4, 5] e suas subnotas. No entanto, por ser uma escala subjetiva, algumas das notas sofrem de clareza e podem ser atribuídas de maneira inconsistente por equipes de enfermagem, dificultando a validação preditiva da ferramenta.

Uma possível alternativa à escala de Braden é utilizar registros eletrônicos em conjunto com aprendizado de máquina, como feito em [6, 7]. Em [7] utilizou-se o banco de dados aberto MIMIC-III [8] para a comparação da escala de Braden com modelos de aprendizado de máquina. Foi reportado que, para uma mesma precisão de 9% a escala de Braden obteve sensibilidade de 50% enquanto um modelo de regressão logística atingiu 71% de sensibilidade.

O objetivo deste artigo é treinar modelos de aprendizado de máquina usando dados de registros eletrônicos da base de dados MIMIC-III e compará-los com a escala de Braden. Além disso, é proposta uma nova abordagem para o problema onde são feitas predições diárias obtendo-se métricas a nível de internação.

## MATERIAIS E MÉTODOS

Foi utilizada a base de dados aberta MIMIC-III v1.4 [8] que possui 61.532 internações de 46.476 pacientes distintos.

### Abordagem

Em [7] a abordagem implementada consiste em utilizar os dados das primeiras 24 horas de internação para predizer uma única vez a incidência de LP. No entanto, a escala de Braden está disponível com maior frequência, geralmente diária. Portanto, o desempenho real da escala de Braden, por envolver uma maior quantidade de informação das múltiplas leituras, será tipicamente superior ao reportado em [7].

Isso motiva a nova abordagem proposta neste artigo, que generaliza a utilizada em [7]. Na nova abordagem considera-se que a escala de Braden será bem sucedida se, em qualquer momento antes da lesão, classificar um paciente que terá LP como em risco. Para pacientes que não terão LP, o acerto será dado se a escala de Braden não classificá-lo como em risco em nenhum momento. Ou seja, o objetivo é aplicar a intervenção apenas nos pacientes que necessitam. Essa abordagem pode ser vista como a agregação das predições diárias de uma internação a partir da operação de OU lógico. A mesma abordagem é utilizada para avaliar os modelos treinados. Por fim, define-se que uma internação é da classe positiva se em qualquer momento ocorreu LP e da classe negativa se não ocorreu LP em nenhum momento.

### Extração de dados e atributos

Para cada dia de internação extraem-se os dados disponíveis tanto nas últimas 24 horas quanto desde o início da internação. No total foram obtidos 40 tipos de informações diferentes referentes aos dados extraídos por [7], incluindo dados de sinais vitais, demográficos, transferências, exames, entre outros. No total, incluindo codificação *one-hot*, temos 80 atributos.

Assim como em [7], são removidos da análise pacientes menores de 18 anos, bem como internações sem nenhuma leitura da escala de Braden. Não são considerados os dias após a incidência de LP estágio 2 ou maior. Após esse processo, se mantiveram 44.366 internações de 33.286 pacientes distintos totalizando 191.134 dias de internação. No total foram registradas 1815 internações com LP nível 2 ou maior.

### Métricas

Foram utilizadas as métricas de sensibilidade e precisão. A sensibilidade, também conhecida como taxa de verdadeiro positivo (TVP), é dada por vp/(vp + fn), enquanto a precisão, também conhecida como valor preditivo positivo (VPP), é dada por vp/(vp + fp), onde vp, fn e fp indicam, respectivamente, o número de verdadeiros positivos, falsos negativos e falsos positivos.

Ambas as métricas estão associadas a grandezas interpretáveis para o nosso problema. A sensibilidade está relacionada à qualidade do atendimento, ou seja, a proporção de pacientes que irão desenvolver uma LP que foram classificados como positivos. Já o inverso da precisão indica, em média, a quantidade de classificações positivas que são necessárias para cada verdadeiro positivo, ou seja, está relacionado ao custo da intervenção. Nota-se que há um *trade-off* entre as duas métricas; portanto, para realizar comparações entre modelos, uma das métricas deve ser fixada.

### Pré-processamento

Cada atributo foi classificado entre categórico, booleano e numérico onde cada uma dessas categorias passa por um tipo de pré-processamento. Atributos numéricos são imputados usando a mediana e em seguida são escalonados subtraindo-se a mediana e dividindo-se pelo intervalo interquartil. Os atributos booleanos foram criados de modo que ao ser imputado com *False* introduz-se a informação esperada ao não ter dados. Os atributos categóricos são imputados com *None*, uma categoria representando a ausência de dados, e por fim é feito o processo de codificação *one-hot*.

O pré-processador é acoplado ao modelo de modo que é ajustado apenas no conjunto de treinamento e as transformações são aplicadas automaticamente na etapa de inferência.

### Treinamento e otimização de hiperparâmetros

No treinamento, em vez de predizer LP em qualquer momento da internação, os modelos são treinados para predizer apenas a incidência de LP nos próximos 7 dias. Isso é feito para evitar que os modelos sofram de sobreajuste tentando predizer uma condição que ainda não apresentou sinais suficientes de que irá ocorrer.

Foi feita uma separação, de maneira estratificada por internações com e sem LP, de conjuntos de treinamento e teste com proporção 80%/20%. No conjunto de treinamento foi feita a otimização dos hiperparâmetros utilizando 30 amostras aleatórias de hiperparâmetros e selecionando a que obteve o melhor desempenho na validação cruzada com 5-folds. A métrica utilizada foi a precisão fixando a sensibilidade em 50%, ou seja, encontra-se o limiar que garante sensibilidade 50% e calcula-se a precisão correspondente. A escala de Braden foi avaliada da mesma maneira utilizando apenas a nota mais recente disponível.

# RESULTADOS

Foram escolhidos cinco modelos de aprendizado de máquina que apresentaram bom desempenho em trabalhos anteriores [7]. Os modelos foram treinados e comparados com a escala de Braden na métrica de precisão para sensibilidade em 50%, conforme mostrado na Tabela 1. Nessa métrica, a escala de Braden obteve precisão de 12,81%, enquanto o melhor modelo obteve 20,99%, o que pode ser interpretado como uma redução de até 40% no custo da intervenção mantendo o mesmo nível de atendimento. A Tabela 1 também mostra a sensibilidade alcançada pelos modelos para a mesma precisão obtida pela escala de Braden (12,81%), indicando que, para um mesmo custo, seria possível melhorar o nível de atendimento em até 64% se usado o melhor modelo encontrado. Em relação aos resultados de [7], nota-se que o desempenho da escala de Braden melhorou significativamente. Além disso, tanto a sensibilidade quanto a precisão dos modelos treinados foram superiores ao melhor modelo de [7].

**Tabela 1 – Métricas de interesse no conjunto de teste**

| Modelo | Precisão para sensibilidade em 50% | Sensibilidade para precisão em 12,81% |
|---|---|---|
| Escala de Braden | 12,81% | 50% |
| Regressão Logística | 20,66% | 80,17% |
| Floresta Aleatória | 19,10% | 80,44% |
| *Gradient Boosting* | 20,52% | 80,99% |
| *XGBoost* | 19,65% | 80,44% |
| *CatBoost* | **20,99%** | **81,82%** |

A Figura 1 apresenta um gráfico da precisão em função da sensibilidade para os modelos treinados e a escala de Braden. Nota-se que os modelos superam a escala de Braden em todos os pontos de operação.

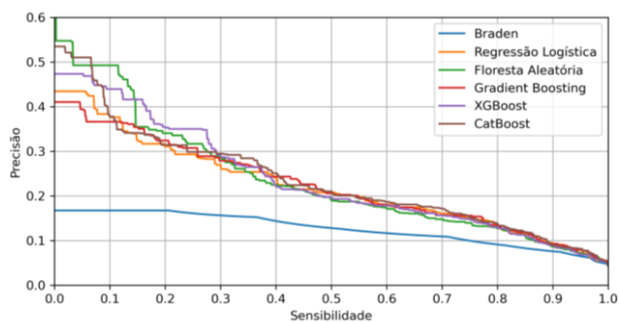

**Figura 1 – Valores de precisão em função da sensibilidade para todos os modelos**

# CONCLUSÃO

Neste artigo é proposta uma nova abordagem para avaliação de ferramentas de predição de LP. A partir dela, foram treinados modelos de aprendizado de máquina e verificou-se que superam o desempenho da escala de Braden. Esse resultado motiva o desenvolvimento de uma ferramenta automática para identificação de pacientes em risco podendo eliminar a mão de obra associada às técnicas utilizadas atualmente, além de promover uma redução no custo da intervenção ou melhoria no nível de atendimento.

Como trabalho futuro, visamos repetir este estudo em uma base de dados de um hospital privado brasileiro.